\documentclass[a4paper,twoside]{article}

\usepackage{epsfig}
\usepackage{subcaption}
\usepackage{calc}
\usepackage{amssymb}
\usepackage{amstext}
\usepackage{amsmath}
\usepackage{amsthm}
\usepackage{multicol}
\usepackage{pslatex}
\usepackage{apalike}
\usepackage{booktabs}
\usepackage{algorithm2e}
\usepackage[bottom]{footmisc}
\usepackage{SCITEPRESS}     % Please add other packages that you may need BEFORE the SCITEPRESS.sty package.
\usepackage{pifont}
\newcommand{\cmark}{\ding{51}}
\newcommand{\xmark}{\ding{55}}
\usepackage{multirow}
\usepackage{makecell}
\usepackage[table]{xcolor}
\usepackage{url}

\begin{document}

\title{ThreatCore: A Benchmark for Explicit and Implicit Threat Detection\thanks{This paper contains examples of potentially offensive or sensitive content. Such material is included solely for scientific research and analysis purposes and does not reflect the authors' views.}}

\author{\authorname{Davide Bruni\sup{1,2}\orcidAuthor{0009-0001-3291-403X}, Carlo Bardazzi\sup{1}\orcidAuthor{0009-0005-5699-5208} and Maurizio Tesconi\sup{2}\orcidAuthor{0000-0001-8228-7807}}
\affiliation{\sup{1}Computer Science Department, University of Pisa, Italy}
\affiliation{\sup{2}
Institute of Informatics and Telematics, National Research Council, Italy}
\email{davide.bruni@phd.unipi.it, c.bardazzi1@studenti.unipi.it, maurizio.tesconi@iit.cnr.it
}
}

\abstract{Threat detection in Natural Language Processing lacks consistent definitions and standardized benchmarks, and is often conflated with broader phenomena such as toxicity, hate speech, or offensive language. In this work, we introduce ThreatCore, a public available benchmark dataset for fine-grained threat detection that distinguishes between explicit threats, implicit threats, and non-threats.
The dataset is constructed by aggregating multiple publicly available resources and systematically re-annotating them under a unified operational definition of threat, revealing substantial inconsistencies across existing labels. To improve the coverage of underrepresented cases, particularly implicit threats, we further augment the dataset with synthetic examples, which are manually validated using the same annotation protocol adopted for the re-annotation of the public datasets, ensuring consistency across all data sources.
We evaluate Perspective API, zero-shot classifiers, and recent language models on ThreatCore, showing that implicit threats remain substantially harder to detect than explicit ones. Our results also indicate that incorporating Semantic Role Labeling as an intermediate representation can improve performance by making the structure of harmful intent more explicit. Overall, ThreatCore provides a more consistent benchmark for studying fine-grained threat detection and highlights the challenges that current models still face in identifying indirect expressions of harmful intent.}

\onecolumn \maketitle \normalsize 

\setcounter{footnote}{0} \vfill

\section{\uppercase{Introduction}}
\label{sec:introduction}
The rapid growth of social media platforms and digital communication channels has expanded the scale at which harmful content can spread online~\cite{zhang2019hate}. Prior work has shown that anonymity, physical distance, and the online disinhibition effect can facilitate hostile and aggressive behavior in digital environments~\cite{bazarova2013managing}. As a result, online platforms have become a fertile ground for multiple forms of textual abuse, ranging from offensive language and hate speech to harassment and direct incitement to violence. Empirical evidence also suggests that online environments may increase exposure to extremist narratives and echo chambers, potentially contributing to violent action tendencies~\cite{molmen2023mechanisms}.

Despite substantial advances in Natural Language Processing (NLP), threat detection remains a challenging task with inconsistent definitions and a lack of standardized benchmarks. In many studies, threatening language is instead treated as a subcategory of broader phenomena such as hate speech, cyberbullying, or toxicity. This framing often leads to systematic ambiguity: models may conflate offensive or hostile language with actual expressions of harmful intent, failing to distinguish aggression from threats. In addition, the limited body of work explicitly devoted to threat detection is fragmented across heterogeneous domains and definitions, ranging from legal intimidation to sexual, political, or identity-based threats. This conceptual fragmentation limits the development of robust and generalizable systems for identifying threats as a distinct phenomenon.

To address this gap, we introduce \textbf{ThreatCore}\footnote{https://github.com/DavideBruni/ThreatCore}, a comprehensive dataset for threat detection structured around three distinct classes: \textbf{explicit threats, implicit threats, and non-threats}. ThreatCore combines multiple data sources and augments them with carefully curated synthetic examples designed to challenge model sensitivity to nuanced and contextually embedded threats. All instances have been systematically re-annotated according to a unified and rigorous operational definition of threat, ensuring clear boundaries between the three classes and preventing the conflation of hostility with violent intent.

The main contributions of this work are threefold:
\begin{enumerate}
    \item \textbf{Conceptual contribution.}
    We introduce an operational definition of threat that separates harmful intent from broader notions such as offensiveness or toxicity, and distinguishes between explicit and implicit threats, providing a clearer basis for annotation and evaluation.

    \item \textbf{Dataset contribution.}  
    We present \textit{ThreatCore}, a multi-source dataset obtained by aggregating and re-annotating publicly available corpora under a unified labeling framework. The re-annotation process resolves inconsistencies across existing datasets and reveals substantial label noise and conceptual inconsistencies in existing resources, challenging their validity as benchmarks.

    \item \textbf{Empirical contribution.}  
    Through extensive experiments, we show that current models, including Perspective API and state-of-the-art LLMs, struggle significantly with implicit threats, despite performing well on explicit ones. This highlights a critical gap in current approaches and motivates the need for more fine-grained modeling of harmful intent.
    
\end{enumerate}

\section{Related Work}\label{sec:related}

This section reviews prior work on threat detection, focusing on threat definitions and datasets.

\paragraph{Threat definitions.}
The notion of threat varies considerably across the literature. Early computational approaches adopt a narrow definition of violent threats based on explicit lexical cues, requiring the co-presence of an aggressor, a violent verb, and a specific target~\cite{hammer2014detecting}. While this formulation enables rule-based detection, it has been criticized as overly restrictive. More recent work emphasizes indirect or implied threats, which may lack explicit violent predicates while still conveying harmful intent. For instance, some studies define threats as expressions of intent to cause harm, distinguishing between direct and implied threats, the latter often conveyed through metaphor, conditional language, or hypothetical scenarios~\cite{raza2024}. Other studies adopt broader definitions, including legal or symbolic harm, and introduce domain-specific distinctions such as sexist vs non-sexist threats~\cite{kumbale2023bree} or judicial vs non-judicial threats~\cite{ravi2023exploring}. 
Threat labels also appear in broader abusive language taxonomies: OLID~\cite{zampieri2019predicting} includes threats under targeted insults, while ETHOS~\cite{mollas2022ethos} distinguishes violent from non-violent hate speech. Overall, these definitions only partially overlap, leading to inconsistencies in annotation and interpretation.

\paragraph{Datasets.}
Several datasets have been proposed for studying threats in online discourse, although availability is limited, with some resources no longer accessible or never released~\cite{wester2016threat,golbeck2017,alharthi2023target,ravi2023exploring}. Existing datasets also differ in their definitions of threat, complicating comparisons and only partially capturing the diversity of threatening language. Moreover, threat detection is frequently treated as a secondary task within broader hate speech or toxicity datasets. As a result, threat instances are typically scarce. For example, the Latent Hatred dataset contains 666 entries labeled as threats instances out of 6,346 total entries~\cite{elsherief2021latent}, while the Gab Hate Corpus includes 918 threats out of 86,529 records~\cite{kennedy2022introducing}. Although this imbalance reflects real-world distributions, it limits the availability of positive examples and makes it difficult to model implicit threats. Notably, a different perspective is offered by the Jigsaw dataset, which includes a continuous threat score, typically binarized at 0.7~\cite{alvisi2025mapping,bruni2025amaqa}, an approach we also adopt.

Moreover, most datasets are derived from social media platforms such as Twitter, Reddit, and YouTube~\cite{hammer2019}, with exceptions such as Wikipedia Talk Pages (Jigsaw) and synthetically generated data~\cite{vidgen2021learning}.

\section{ThreatCore Dataset}
\subsection{Threat Definition and Label Schema}\label{subsec:def}
Given the class imbalance of available resources and the heterogeneity of threat definitions discussed in Section~\ref{sec:related}, we first establish a unified operational definition of \textit{threat} to guide dataset construction and annotation.
We adopt the following definition:

\begin{quote}
Threat: A statement or phrase intended to announce harm, injury, or punishment to a target, primarily serving to intimidate or coerce the target into performing or refraining from a specific action.
\end{quote}

Our definition of threat is grounded in prior literature. In particular, we build on definitions such as the one proposed in BREE-HD~\cite{kumbale2023bree}, which describe threats as statements expressing an intention to inflict harm.
We adopt a compatible but more operational formulation, defining a threat as a statement intended to communicate harm, injury, or punishment toward a target, typically with the purpose of intimidation or coercion. This definition is further aligned with lexical definitions of threat in general-purpose language resources\footnote{Threat definition from the Italian dictionary Dizionario Internazionale (“minaccia” entry), available at: https://dizionario.internazionale.it/parola/minaccia}. Unlike broader definitions used in hate speech detection, our formulation requires the presence of intent to cause harm, which is necessary to distinguish threats from merely offensive or aggressive language. This distinction is crucial for annotation consistency and enables a clearer separation between threat and non-threat categories across heterogeneous data sources.

Moreover, we introduced a finer-grained distinction between \textit{explicit} and \textit{implicit} threats. As a result, the ThreatCore dataset comprises the following three classes:

\begin{itemize}
    \item \textbf{Explicit Threat}: 
    Text that explicitly expresses an intent to inflict harm upon an entity (individual, organization, country, etc.). The presence of vulgarity, swearing, or profanity is not a necessary condition for this classification.
    
    \item \textbf{Implicit Threat}:
    Text that implicitly conveys an intent to inflict harm upon an entity. These instances lack direct violent vocabulary (e.g., ``kill'') and may rely on external context, metaphors, or seemingly non-aggressive tones to deliver the warning.
    
    \item \textbf{Non-Threat}: 
    A message devoid of any intent to threaten harm. This category includes texts that may be vulgar, offensive, or violent in nature (e.g., hate speech without a threat), as well as entirely neutral or polite messages.
\end{itemize}

\subsection{Source Datasets and Initial Collection}
To build the \textbf{ThreatCore} dataset, we aggregated publicly available datasets into a unified corpus. After removing duplicates, we obtained 19,563 potentially relevant instances.
ThreatCore combines the publicly available and accessible threat-related datasets that we were able to retrieve and validate for this study. Datasets such as  Multi-Level-Threats \cite{ravi2023exploring}, Target-Oriented Abuse \cite{alharthi2023target}, Large Labeled Corpus \cite{golbeck2017}, AMI-Dataset \cite{anzovino2018automatic}, and THREAT \cite{hammer2019} were excluded due to unavailability, as their original publications provided missing or non-functional access links. The only exception is OLID, which was deliberately excluded. As shown in Table~\ref{tab:public_dataset_new-map}, after the re-annotation phase (Section~\ref{subsec:annotation}) only 21 out of 4,088 instances (less than 1\%) were labeled as actual threats. Including it would have introduced strong imbalance while contributing negligible signal, so it was omitted from the final dataset.

\begin{table*}
\caption{Publicly available threat datasets. \textbf{Source} indicates the original data source, and \textbf{Tot} the total number of records. \textbf{Label} and \textbf{Threat} refer to the original labels and their counts. \textbf{Exp}, \textbf{Imp}, and \textbf{Non} denote explicit, implicit, and non-threat instances after re-annotation process. \textbf{Incl} denotes whether the dataset is included in the final ThreatCore corpus.}
\label{tab:public_dataset_new-map}\centering
\begin{tabular}{l l r l r r r r c}
\toprule

& & \multicolumn{3}{c}{Origin} & \multicolumn{3}{c}{ThreatCore} & \\

\cmidrule(lr){3-5} \cmidrule(lr){6-8}

Dataset & Source & Tot & Label & Threat & Exp & Imp & Non & Incl \\

\midrule

Jigsaw Unintended Bias & Wikipedia & 1.7M & Threat  & 1457 & 1076 & 56 & 325 & \cmark \\
\midrule

\makecell[l]{Latent Hatred \\ \cite{elsherief2021latent}} & Twitter & 6346 & Threatening & 666 & 149 & 64 & 453 & \cmark \\
\midrule

\makecell[l]{DynHate\\ \cite{vidgen2021learning}} & Synthetic & 41144 & Threatening & 606 & 476 & 47 & 83 & \cmark \\
\midrule

\makecell[l]{Gab Hate Corpus \\ \cite{kennedy2022introducing}} & Gab & 86529 & Threat & 918 & 329 & 88 & 494 & \cmark \\
\midrule

\makecell[l]{ETHOS \\ \cite{mollas2022ethos}} & \makecell[l]{Youtube \\ Reddit} & 433 & Violence & 142 & 108 & 11 & 23 & \cmark \\
\midrule

\multirow{2}{*}{\makecell[l]{BRET-HD \\ \cite{kumbale2023bree}}}
& \multirow{2}{*}{Twitter} 
& \multirow{2}{*}{5888} 
& sexist-threat & 900 & 818 & 72 & 10 & \multirow{2}{*}{\cmark} \\
& & & non-sexist-threat & 1193 & 83 & 10 & 1100 & \\
\midrule

\multirow{2}{*}{\makecell[l]{ThreatGram101 \\ \cite{ravi2024threatgram101}}}
& \multirow{2}{*}{Telegram} 
& \multirow{2}{*}{15076} 
& judicial-threat & 4658 & 90 & 58 & 4510 & \multirow{2}{*}{\cmark} \\
& & & non-judicial-threat & 5393 & 3206 & 403 & 1784 & \\
\midrule

\makecell[l]{OLID \\ \cite{zampieri2019predicting}} & Twitter & 13241 & Targeted Insult & 4088 & 15 & 6 & 4067 & \xmark \\

\bottomrule
\end{tabular}
\end{table*}

\subsection{Re-annotation Protocol}\label{subsec:annotation}
A preliminary sample-based analysis of the collected datasets revealed that many instances originally labeled as \textit{threat} did not align with the definition adopted in this work ( Section~\ref{subsec:def}). This discrepancy motivated a full re-annotation process aimed at improving conceptual consistency across sources.

We adopted a double-coding protocol in which two annotators independently labeled all candidate instances. The annotation was carried out by two English-fluent annotators, a PhD candidate and a Master's degree student, following a shared set of written guidelines. The goal of the process was not to preserve original labels, but to re-assess each instance according to a unified notion of threat and the explicit/implicit/non-threat distinction.

Inter-annotator agreement, measured with Cohen’s kappa, indicates near-perfect agreement ($\kappa = 0.91$), supporting the clarity of the annotation framework.

\subsection{Re-annotation Outcomes and Dataset-Level Observations}
\begin{figure}[h]
    \centering
    \includegraphics[width=1\linewidth]{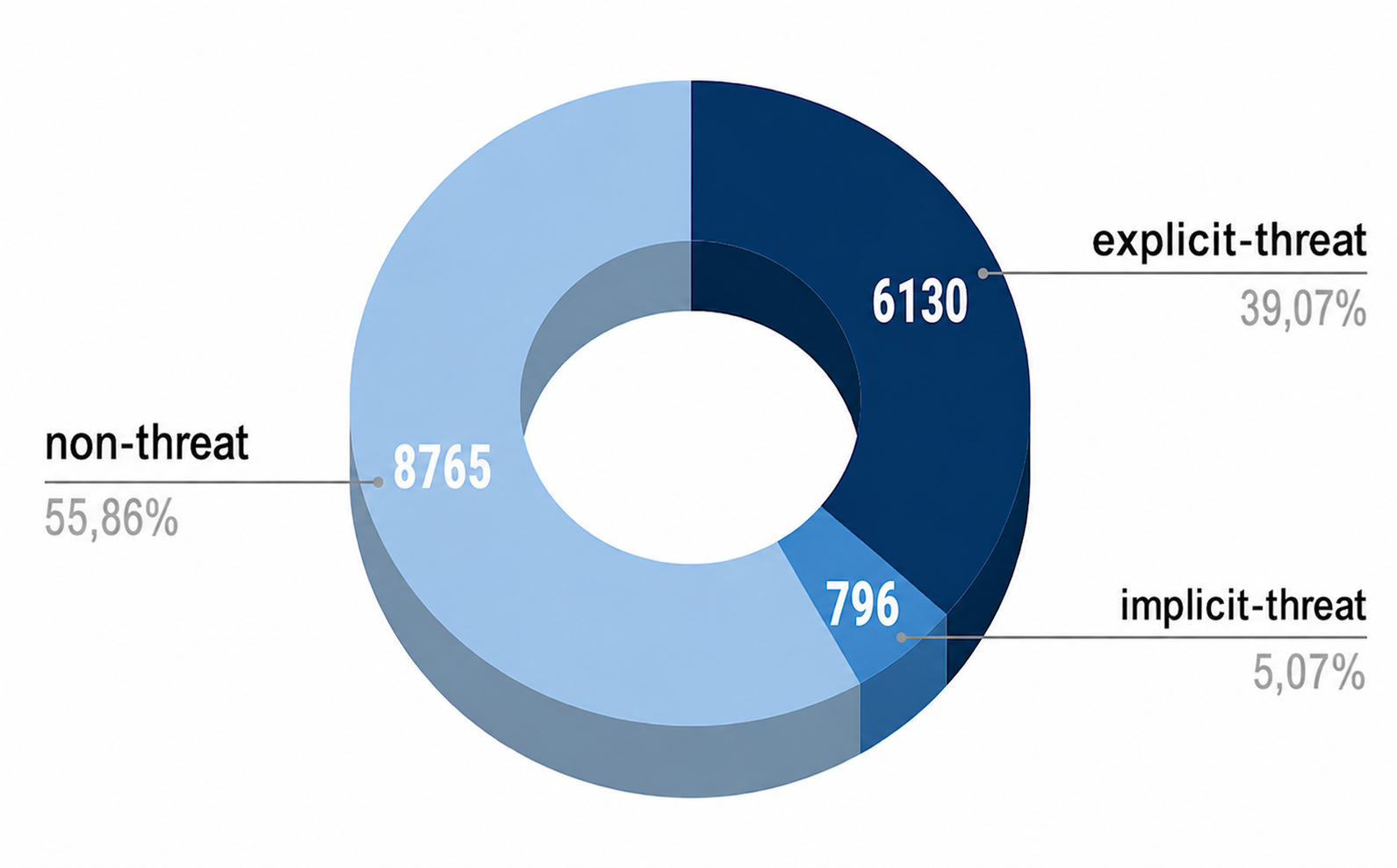}
    \caption{Class distribution after re-annotation phase.}
    \label{fig:reannotation-distribution}
\end{figure}

After the re-annotation process, the resulting ThreatCore dataset comprises 15,691 instances, with a predominance of non-threat instances (55,86\%), followed by explicit threats (39.07\%) and a smaller proportion of implicit threats (5.07\%), as illustrated in Figure~\ref{fig:reannotation-distribution}. This redistribution reveals a clear misalignment between the original labels and our definition of threat, highlighting substantial label noise and conceptual inconsistencies, as many instances originally labeled as threats reflect broader categories such as toxicity, hate speech, or offensive language and are therefore reclassified as non-threat.

Table~\ref{tab:public_dataset_new-map} provides a detailed view of this redistribution by reporting, for each original dataset, the mapping between original and ThreatCore labels, along with the data source, total number of records, and inclusion status in the final collection.

Beyond these quantitative aspects, several qualitative patterns emerge from the re-annotation process. First, datasets originally labeled with generic or loosely defined threat categories (e.g., \textit{Threat}, \textit{Threatening}) exhibit substantial variability in how threatening content is operationalized. For instance, in Jigsaw Unintended Bias and Gab Hate Corpus, only a fraction of the originally labeled threat instances are confirmed as explicit or implicit threats, with a non-negligible portion reclassified as non-threat. This suggests that these datasets often conflate threats with broader categories such as toxicity, hate speech, or offensive language.
More dataset-specific patterns further highlight annotation inconsistencies. 
In BRET-HD, the \textit{non-sexist-threats} are largely reclassified as non-threat, as it predominantly contains insulting or abusive language without a clear expression of intent to cause harm. In contrast, the \textit{sexist-threats} more often contains violent expressions and are frequently labeled as explicit threats.

In ThreatGram101, the \textit{judicial-threat} label is mostly reclassified as non-threat, as it often includes slogan-like or politically charged expressions (e.g., “Lock her up!”) that lack explicit intent to harm or coerce a target, and therefore do not meet our criteria for either explicit or implicit threats.

Overall, these observations underline the limitations of existing datasets, where threat-related labels are often entangled with broader notions of offensiveness, hostility, or punitive rhetoric.

\subsection{Dataset Augmentation}
As shown in Figure~\ref{fig:reannotation-distribution}, the dataset resulting from the re-annotation phase is strongly imbalanced, with a strong predominance of non-threat instances and very few implicit threats. This is particularly problematic, as implicit threats are both rare and difficult to detect, limiting the reliability of evaluation metrics and hindering a meaningful assessment of model performance on this category.

To mitigate this issue, we augmented the dataset with synthetic data generated via LLMs, producing 2,000 explicit and 4,073 implicit threats. The generation process was designed to produce plausible and contextually coherent examples, rather than arbitrary or templated data. Within the implicit threats, we also created a subset referencing real-world events (e.g., crimes, accidents, natural disasters) that occurred after the models’ knowledge cutoff, ensuring that models could not rely on memorized cases but instead had to infer threatening intent from linguistic cues.

For the generation process, we employed Grok-3 through the Azure API. However, due to content safety guardrails, particularly restrictive for explicit threats, we generated explicit threat instances locally using the Qwen-2.5:7B model via Ollama.

All synthetic instances were then manually re-annotated by the original annotators following the same protocol, achieving near-perfect agreement (Cohen’s Kappa $k = 0.964$), confirming the reliability of the generated samples.

\begin{figure}
    \centering
    \includegraphics[width=1\linewidth]{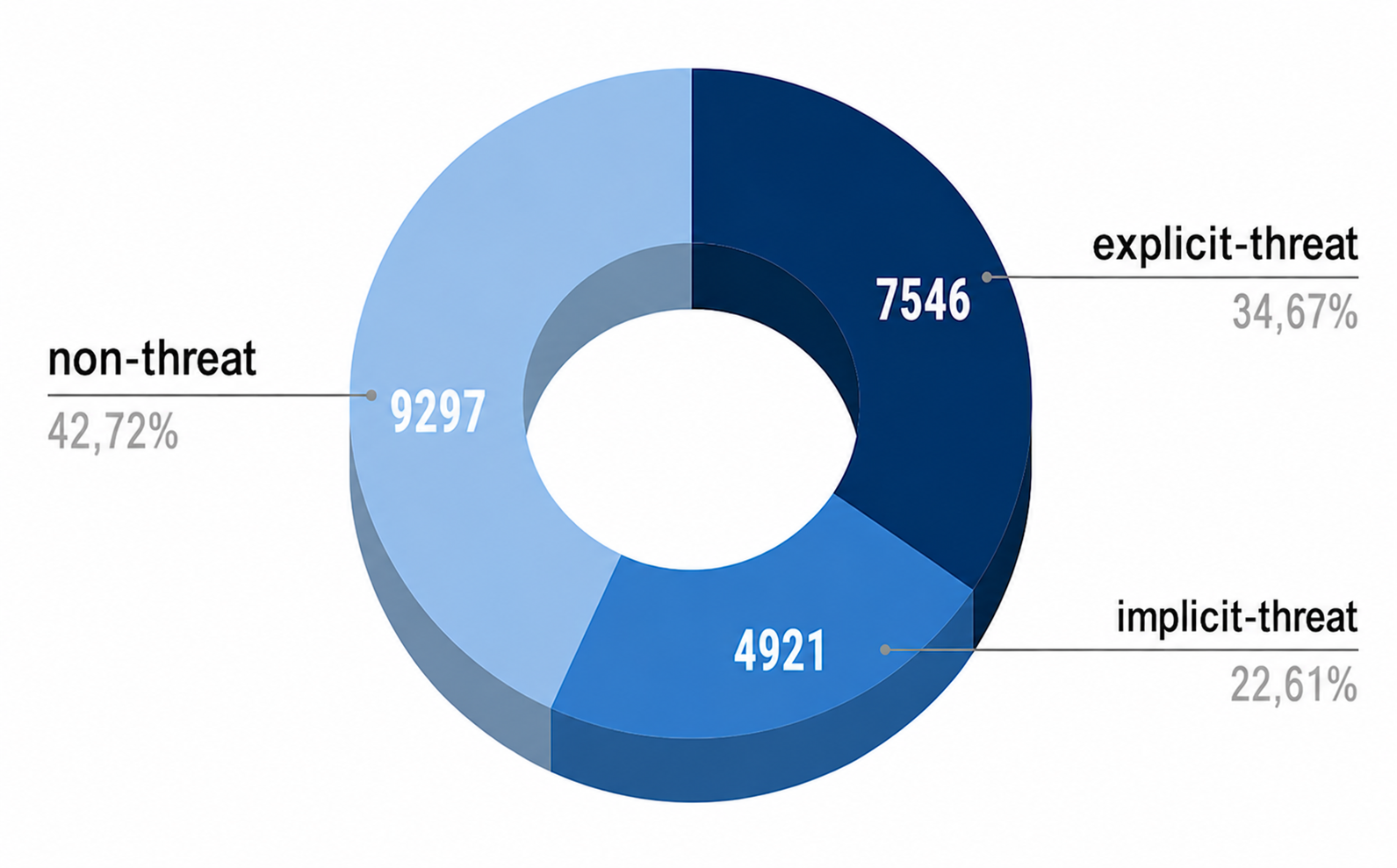}
    \caption{Class distribution of the ThreatCore dataset after the data augmentation process.}
    \label{fig:synth_reannotation-distribution}
\end{figure}

After validation, the final synthetic subset consists of 1,416 explicit threats, 4,125 implicit threats, and 532 non-threat instances. Overall, the augmentation process results in a more balanced and representative dataset, enabling a more robust evaluation of models, particularly for implicit threats. The final ThreatCore dataset comprises 21,764 instances, distributed as shown in Figure~\ref{fig:synth_reannotation-distribution}.

\section{Experiments}
This section describes the experimental setup and presents the evaluation results. All models were evaluated using per-class Precision, Recall, and F1-score, along with Accuracy and Macro F1. Each experiment was run five times, and results are reported as mean performance. When open LLMs were used, they were deployed locally on Ollama using two NVIDIA RTX 6000 Ada GPUs (48 GB VRAM each).

\subsection{Perspective API}
We first evaluate Perspective API\footnote{https://perspectiveapi.com/}, a widely adopted industrial system for toxicity and threat detection. The API outputs a continuous threat score in the range [0,1] without distinguishing between implicit and explicit threats. Following prior works, we apply a threshold of 0.7 to obtain binary threat predictions~\cite{alvisi2025mapping,bruni2025amaqa}.

\begin{table}[h]
    \centering
    \caption{Performance Metrics for Perspective API.}
    \vspace{0.2cm}
    \begin{tabular}{lccc}
        \toprule
        Class & Precision & Recall & F1-Score \\
        \midrule
        Explicit Threat & 0.962 & 0.221 & 0.360 \\
        Implicit Threat  & 0.283 & 0.005 & 0.010 \\
        Non-Threat       & 0.462 & 0.993 & 0.630 \\
        \midrule
        Macro-F1 & \multicolumn{3}{c}{0.333} \\
        \bottomrule
    \end{tabular}
    \label{tab:perspective_api_results}
\end{table}
As shown in Table~\ref{tab:perspective_api_results}, Perspective API achieves extremely high precision on explicit threats, but with very low recall, indicating a strongly conservative behavior. The recall on implicit threats is close to zero, suggesting a very limited ability to capture indirect or context-dependent harmful intent. By contrast, the model achieves near-perfect recall on non-threat content, pointing to a strong bias toward the safe class, consistent with its design as a general-purpose moderation tool.
Since the API is also a black-box system that cannot be adapted or fine-tuned for this task, its usefulness in our setting appears limited.

\subsection{Zero-Shot Classifier}
We evaluate the ModernBERT-base-zeroshot-v2.0 model ~\cite{laurer_modernbert_zeroshot_2024}, which represents the current state of the art in zero-shot textual classification. The model is evaluated using the hypothesis template: ``This text belongs to class \{\}'' together with the names of the three target classes.

\begin{table}[h]
    \centering
    \caption{Performance Metrics for ModernBERT-base-v2.0.}
    \vspace{0.2cm}
    \begin{tabular}{lccc}
        \toprule
        Class & Precision & Recall & F1-Score \\
        \midrule
        Explicit Threat & 0.427 & 0.402 & 0.414 \\
        Implicit Threat & 0.328 & 0.901 & 0.481 \\
        Non-Threat      & 0.744 & 0.092 & 0.164 \\
        \midrule
        Macro-F1 & \multicolumn{3}{c}{0.502} \\
        \bottomrule
    \end{tabular}
    \label{tab:modernbert_zeroshot_results}
\end{table}

As shown in Table~\ref{tab:modernbert_zeroshot_results}, the zero-shot configuration exhibits a strong tendency to over-predict threat-related categories, particularly implicit threats. This is reflected in the very high recall achieved on implicit threats, coupled with low precision, indicating that many non-implicit instances are incorrectly assigned to that class. Explicit threats display a more balanced but still limited performance, with both precision and recall remaining moderate.

The most critical issue concerns the non-threat class, whose recall drops close to zero despite relatively high precision. This suggests that the model rarely predicts non-threat content, favoring threat labels even in ambiguous cases. While this behavior may reduce missed threats, it limits the practical usability of the model due to the high number of false positives.

\subsection{Large Language and Reasoning Models}
We evaluate a set of proprietary and open-weight Large Language Models (LLMs) and Large Reasoning Models (LRMs), including gpt-4o-mini, grok3, qwen2.5:7b, llama3.3:70b, and phi4:14b. Larger models were excluded due to hardware constraints. Prompts used in the experiments are available in the public repository.

For each model, we tested five prompt configurations, including zero-shot and few-shot settings, varying the instruction structure. Table~\ref{tab:best_models_comparison} reports the results for the best-performing configuration of each model.

\begin{table*}
    \caption{Class-wise performance of evaluated LLM and LRM models. Precision (P), Recall (R), and F1 are reported per threat category, along with macro-F1. The best value for each metric is highlighted in yellow.}
    \centering
    \begin{tabular}{l|ccc|ccc|ccc|c}
        \toprule
        & \multicolumn{3}{c|}{Non-Threat} & \multicolumn{3}{c|}{Explicit Threat} & \multicolumn{3}{c|}{Implicit Threat} & Global\\
        Model 
        & P & R & F1
        & P & R & F1 
        & P & R & F1 
        & Macro-F1 \\
        \midrule
        GPT-oss:20b & 0.864 & 0.658 & 0.747 & 0.674 & \cellcolor{yellow!30}0.770 & 0.719 & 0.632 & 0.777 & \cellcolor{yellow!30}0.697 & \cellcolor{yellow!30}0.721 \\
        GPT-4o-mini & 0.903 & 0.562 & 0.693 & 0.750 & 0.737 & \cellcolor{yellow!30}0.743 & 0.482 & 0.839 & 0.613 & 0.683 \\
        Grok-3 & 0.722 & 0.833 & 0.773 & 0.651 & 0.568 & 0.606 & \cellcolor{yellow!30}0.672 & 0.607 & 0.638 & 0.673 \\
        Phi4:14b & 0.904 & \cellcolor{yellow!30}0.858 & \cellcolor{yellow!30}0.880 & 0.736 & 0.734 & 0.734 & 0.337 & 0.456 & 0.375 & 0.663 \\
        Qwen2.5:7b & 0.757 & 0.476 & 0.584 & 0.599 & 0.699 & 0.645 & 0.513 & 0.741 & 0.607 & 0.612 \\
        Llama3.3:70b & \cellcolor{yellow!30}0.927 & 0.322 & 0.478 & \cellcolor{yellow!30}0.756 & 0.651 & 0.700 & 0.364 & \cellcolor{yellow!30}0.891 & 0.517 & 0.565 \\
        \bottomrule
    \end{tabular}
    \label{tab:best_models_comparison}
\end{table*}

As shown in Table~\ref{tab:best_models_comparison}, GPT-oss:20b achieves the highest Macro-F1, indicating the most balanced overall behavior across categories. Although it does not consistently obtain the best score for every individual metric, it maintains strong performance on both explicit and implicit threats while avoiding the extreme biases observed in other models. 

Different models exhibit markedly different prediction tendencies. Llama3.3:70b, for instance, adopts a highly conservative strategy on the Non-Threat class: it achieves the highest precision, but with very low recall, meaning that many benign instances are incorrectly flagged as threatening. At the same time, the model captures most implicit threats, suggesting a strong preference toward threat-sensitive predictions that comes at the cost of numerous false positives.

Phi4:14b displays the opposite behavior. It performs particularly well on Non-Threat detection, achieving the highest F1 score for this class, and also maintains solid performance on Explicit Threats. However, it struggles substantially on Implicit Threats, where both precision and recall remain low. Results suggests that the model is effective at recognizing safe content and direct harmful intent, but has difficulty identifying more subtle forms of harmful language, which are inherently harder to detect.

Beyond quantitative performance, LLMs also provide explanations for their predictions when properly prompted. In particular, they can justify why a message is classified as an explicit threat, implicit threat, or non-threat by referring to linguistic and contextual cues, for example distinguishing genuinely threatening content from messages that are merely insulting, hostile, or toxic. Examples of model explanations across different threat categories are provided in the public repository.

\subsection{Semantic Role Labeling}
\begin{figure}[h]
    \centering
    \includegraphics[width=1\linewidth]{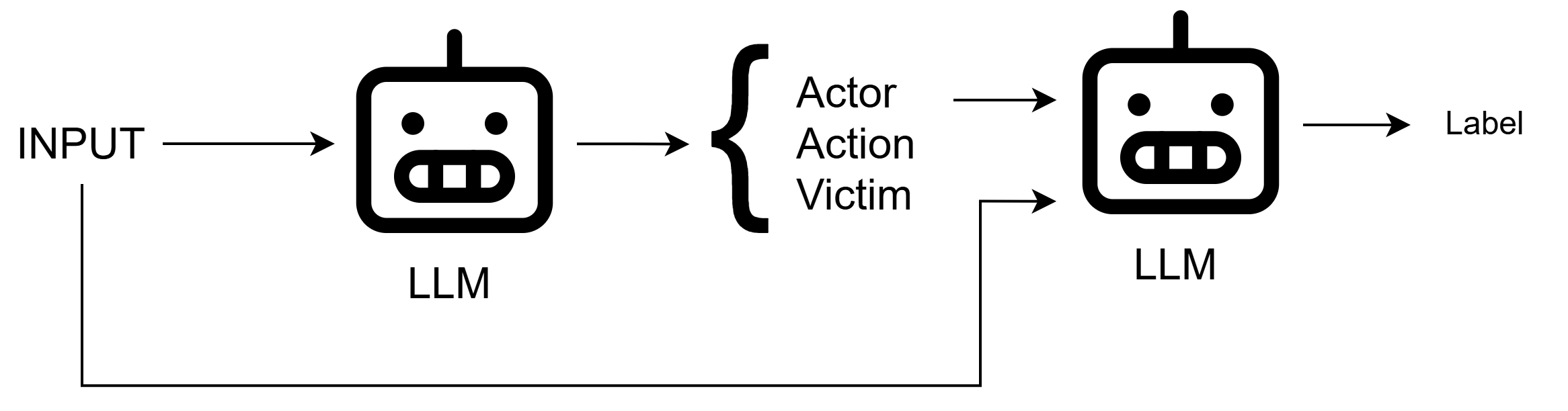}
    \caption{Process of using Semantic Role Labeling with LLMs.}
    \label{fig:schema_srl}
\end{figure}

The goal of semantic role labeling (SRL) is to identify and label the arguments of semantic predicates in a sentence according to predefined relations, such as “who” did “what” to “whom”~\cite{roth2015context}. We leverage this idea to improve threat detection by prompting an LLM to explicitly extract three semantic roles relevant to harmful interactions: \textit{Actor}, defined as the semantic agent who will cause or inflict the harm; \textit{Action}, representing the specific act that causes harm; and \textit{Victim}, the entity receiving the harm or damage (e.g., a person, group of people, or animal). These extracted roles are then provided together with the original text within the prompt for the LLM used as classifier. The full process is depicted in Figure~\ref{fig:schema_srl}.

As shown in Table~\ref{tab:framework_v2_results_updated}, the inclusion of structured semantic information through SRL generally improves performance across the evaluated models, while also altering their relative ranking. \textit{Qwen2.5:7b} achieves the highest precision in the Non-Threat class, whereas \textit{Llama3.3:70b} obtains more balanced but overall weaker results.

The largest gains are observed for \textit{Phi4:14b}, which achieves the best F1 scores across all three classes as well as the highest Macro-F1, outperforming not only the other SRL-based configurations but also the best-performing model without SRL (\textit{GPT-oss:20b}). In particular, the model attains the highest recall for both Non-Threat and Implicit Threat, suggesting that SRL substantially improves its ability to capture subtle and context-dependent forms of harmful language.

Overall, these findings indicate that explicitly modeling semantic roles enhances the understanding of threatening language, especially for implicit threats, where harmful intent must be inferred rather than explicitly stated. However, the impact is not uniform across models: while \textit{Phi4:14b} benefits considerably from the additional semantic structure, others, such as \textit{GPT-oss:20b}, show limited or even slightly degraded performance. This suggests that the effectiveness of SRL is strongly model-dependent and related to how different architectures exploit structured semantic information.

Finally, GPT-4o-mini and Grok-3 could not be evaluated within the SRL-based framework. Both models were accessed through Azure APIs whose safety mechanisms prevented the generation of violent or harmful content. Since SRL approach requires explicit extraction of the harmful \textit{Action}, these restrictions made evaluation infeasible. 

\begin{table*}
    \caption{Class-wise performance of the evaluated models using Semantic Role Labels within the prompt. Precision (P), Recall (R), and F1 are reported per threat category, along with macro-F1. The best value for each metric is highlighted in yellow.}
    \centering
    \label{tab:framework_v2_results_updated}
        \begin{tabular}{l|ccc|ccc|ccc|c}
            \toprule
            & \multicolumn{3}{c|}{Non-Threat} & \multicolumn{3}{c|}{Explicit Threat} & \multicolumn{3}{c|}{Implicit Threat} & Global \\
            Model 
            & P & R & F1
            & P & R & F1 
            & P & R & F1 
            & Macro-F1 \\
            \midrule
            Phi4:14b     & 0.831 & \cellcolor{yellow!30}0.795 & \cellcolor{yellow!30}0.812 & \cellcolor{yellow!30}0.865 & 0.674 & \cellcolor{yellow!30}0.757 & 0.640 & \cellcolor{yellow!30}0.906 & \cellcolor{yellow!30}0.749 & \cellcolor{yellow!30}0.773 \\
            Qwen2.5:7b   & \cellcolor{yellow!30}0.865 & 0.575 & 0.691 & 0.652 & \cellcolor{yellow!30}0.766 & 0.704 & 0.618 & 0.843 & 0.713 & 0.703 \\
            GPT-oss:20b      & 0.786 & 0.683 & 0.729 & 0.599 & 0.760 & 0.668 & \cellcolor{yellow!30}0.743 & 0.605 & 0.664 & 0.687 \\
            Llama3.3:70b  & 0.840 & 0.629 & 0.719 & 0.672 & 0.727 & 0.699 & 0.509 & 0.687 & 0.585 & 0.668 \\
            \bottomrule
        \end{tabular}%
\end{table*}

\subsection{Anecdotal error analysis}

In this section, we provide a qualitative analysis of model errors to better understand the limitations of the systems. Unlike quantitative metrics, this approach examines specific failure cases, highlighting issues such as classification ambiguity, and sensitivity to superficial textual variations.

\paragraph{LLM and LRM}
The qualitative analysis of errors reveals several recurring patterns.
First, models tend to overestimate threats, misclassifying hostile or punitive statements (e.g., “All the FBI should be put in prison”, “lock her up!!”) as threats (explicit or implicit) despite being non-threatening. This suggests an incorrect association between hostility and threat. Second, models struggle to identify threat forms: for example “non binary nnnnnneed to be shhhhhhhhhhotttttt” is not recognized as an explicit threat, likely due to textual noise, which may instead be interpreted as sarcasm or stylistic emphasis rather than a genuine threat. This highlights limited robustness to orthographic variation, common in online content. 

\paragraph{LLM with Semantic Role Labeling (SRL)}
The introduction of SRL partially changes the behavior of the model but does not eliminate certain limitations. In several cases, the model continues to classify non-threatening sentences as implicit threats, such as “Arrest and remove the illegitimate bastard”. This suggests that the model still tends to interpret calls to action or political expressions as signals of threat.

An interesting pattern concerns the reduction of missed detections: genuinely threatening sentences, such as “He needs to be tortured”, are at least recognized as threats, although they are classified as implicit rather than explicit. This suggests that SRL helps the model capture the presence of a violent event (action + patient), but does not always enable it to correctly distinguish the level of explicitness of the threat.

\section{Discussion}
The results show that fine-grained threat detection remains challenging, especially for implicit threats. As summarized in Table~\ref{tab:overall_summary}, general-purpose moderation tools and zero-shot classifiers are not sufficient for this task. Perspective API behaves conservatively, achieving high precision on explicit threats but very low recall, particularly for implicit threats. Conversely, ModernBERT tends to over-predict threat-related classes, leading to many false positives on non-threatening content.

\begin{table}[h]
    \caption{Overall comparison of methods sorted by Macro-F1, reporting Macro-averaged Precision (P), Recall (R), and F1-score.} \centering
    \begin{tabular}{lccc}
        \toprule
        Model \& method & P & R & F1 \\
        \midrule
        Phi4:14b + SRL & 0.779 & 0.792 & 0.773 \\
        GPT-oss:20b & 0.723 & 0.735 & 0.721 \\
        ModernBERT   & 0.500 & 0.465 & 0.502 \\
        Perspective API & 0.569 & 0.406 & 0.333 \\
        \bottomrule
    \end{tabular}
    \label{tab:overall_summary}
\end{table}

LLM-based approaches provide more balanced results. Among models without additional structure, GPT-oss:20b achieves the best Macro-F1, suggesting that instruction-following models can capture threat-related intent more effectively than generic classifiers. However, qualitative errors show that these models still confuse hostility, punitive language, or aggressive political expressions with actual threats, and they do not always distinguish explicit from implicit threats reliably.

The best overall performance is obtained by Phi4:14b with SRL. This suggests that making the Actor, Action, and Victim structure explicit helps models reason about harmful intent, particularly in implicit cases. Nevertheless, the benefit is model-dependent, and SRL does not fully eliminate false positives or boundary errors between explicit and implicit threats.

Overall, the findings support the need for benchmarks such as ThreatCore, where threat detection is evaluated as a distinct task rather than as a subcategory of hate speech. %They also indicate that structured prompting strategies, such as SRL-based representations, are a promising direction for improving robustness and interpretability in threat detection.

\section{Limitations and Future works}
This work presents several limitations that should be acknowledged.

First, the experiments are conducted exclusively on English-language data. This limits the generalizability of the findings to other linguistic and cultural contexts. Threatening language can vary significantly across languages, not only in terms of vocabulary and grammar but also in the way harmful intent is expressed implicitly. As a result, the current framework may not directly transfer to other languages without further validation.

Second, the dataset includes a substantial portion of synthetic data. All generated instances were manually re-annotated, achieving near-perfect inter-annotator agreement, which supports their reliability. Nevertheless, this does not guarantee the absence of bias or that the synthetic samples fully reflect real-world data (e.g., in terms of length, lexical choices, or stylistic variation).

Third, the evaluation of large language models was constrained by computational resources. Larger open-weight models were not tested, and proprietary models were partially limited by Azure API safety policies, preventing a complete assessment.

Future work will focus on extending the dataset to include multiple languages in order to capture both linguistic and cultural variations in the expression of threats. A multilingual dataset would allow for a more comprehensive evaluation of threat detection systems and enable the investigation of cross-lingual generalization capabilities of large language models. In this context, LLMs could further enhance the usability of threat detection systems by providing natural language explanations tailored to end users. Such explanations may be particularly valuable in multilingual scenarios, helping users understand potentially harmful content written in unfamiliar languages, thereby improving transparency and trust in real-world applications.

\section{Conclusion}
In this work, we introduced \textbf{ThreatCore}, a novel dataset for threat detection that distinguishes between explicit threats, implicit threats, and non-threat content. The dataset contains 15,691 instances from existing datasets, re-annotated under a unified annotation framework, and further extended with approximately 6,000 synthetic examples, for a total of 21,764 instances.

Our experimental results highlight clear limitations in existing approaches. General-purpose moderation systems and zero-shot classifiers struggle to capture the complexity of implicit threats, often exhibiting strong biases toward either over-predicting or under-detecting harmful content. In contrast, large language models demonstrate a more balanced understanding of threatening language, even in zero-shot settings.
We further show that incorporating structured semantic information through Semantic Role Labeling could improves performance by making interaction patterns explicit, enabling models to better reason about the underlying intent of a statement. 

In conclusion, the ThreatCore dataset and the evaluation framework introduced in this work aim to support future research on threat detection and to facilitate the development of more robust systems capable of identifying both explicit and implicit manifestations of violent intent. The dataset, the annotation guidelines, and prompts used in this work are available in the public repository, to encourage collaboration and future advancements in this field.

\bibliographystyle{apalike}
{\small
\bibliography{example}}

@article{zhang2019hate,
  author  = {Zhang, Ziqi and Luo, Lei},
  title   = {Hate speech detection: A solved problem? The challenging case of long tail on {Twitter}},
  journal = {Semantic Web},
  volume  = {10},
  number  = {5},
  pages   = {925--945},
  year    = {2019}
}

@article{bazarova2013managing,
  author  = {Bazarova, Natalya N. and others},
  title   = {Managing impressions and relationships on {Facebook}: Self-presentational and relational concerns revealed through the analysis of language style},
  journal = {Journal of Language and Social Psychology},
  volume  = {32},
  number  = {2},
  pages   = {121--141},
  year    = {2013}
}

@article{raza2024,
  title={Reading Between the Lines: Machine Learning Ensemble and Deep Learning for Implied Threat Detection in Textual Data},
  author={Raza, Muhammad Owais and others},
  journal={International Journal of Computational Intelligence Systems},
  volume={17},
  number={1},
  pages={183},
  year={2024}
}

@article{molmen2023mechanisms,
  title={Mechanisms of online radicalisation: how the internet affects the radicalisation of extreme-right lone actor terrorists},
  author={M{\o}lmen, Guri Nordtorp and Ravndal, Jacob Aasland},
  journal={Behavioral Sciences of Terrorism and Political Aggression},
  volume={15},
  number={4},
  pages={463--487},
  year={2023},
  publisher={Taylor \& Francis}
}

@inproceedings{elsherief2021latent,
  title={Latent hatred: A benchmark for understanding implicit hate speech},
  author={ElSherief, Mai and Ziems, Caleb and Muchlinski, David and Anupindi, Vaishnavi and Seybolt, Jordyn and De Choudhury, Munmun and Yang, Diyi},
  booktitle={Proceedings of the 2021 conference on empirical methods in natural language processing},
  pages={345--363},
  year={2021}
}

@article{kennedy2022introducing,
  title={Introducing the gab hate corpus: defining and applying hate-based rhetoric to social media posts at scale},
  author={Kennedy, Brendan and Atari, Mohammad and Davani, Aida Mostafazadeh and Yeh, Leigh and Omrani, Ali and Kim, Yehsong and Coombs Jr, Kris and Havaldar, Shreya and Portillo-Wightman, Gwenyth and Gonzalez, Elaine and others},
  journal={Language Resources and Evaluation},
  volume={56},
  number={1},
  pages={79--108},
  year={2022},
  publisher={Springer}
}

@inproceedings{ravi2024threatgram101,
  title={ThreatGram101: Extreme Telegram Replies Data with Threat Levels},
  author={Ravi, Kamalakkannan and Yuan, Jiann-Shiun},
  booktitle={Annual International Conference on Information Management and Big Data},
  pages={275--291},
  year={2024},
  organization={Springer}
}

@article{kumbale2023bree,
  title={BREE-HD: A transformer-based model to identify threats on Twitter},
  author={Kumbale, Sinchana and Singh, Smriti and Poornalatha, G and Singh, Sanjay},
  journal={IEEE Access},
  volume={11},
  pages={67180--67190},
  year={2023},
  publisher={IEEE}
}

@inproceedings{zampieri2019predicting,
  title={Predicting the type and target of offensive posts in social media},
  author={Zampieri, Marcos and Malmasi, Shervin and Nakov, Preslav and Rosenthal, Sara and Farra, Noura and Kumar, Ritesh},
  booktitle={Proceedings of the 2019 Conference of the North American Chapter of the Association for Computational Linguistics: Human Language Technologies, Volume 1 (Long and Short Papers)},
  pages={1415--1420},
  year={2019}
}

@article{mollas2022ethos,
  title={ETHOS: a multi-label hate speech detection dataset},
  author={Mollas, Ioannis and Chrysopoulou, Zoe and Karlos, Stamatis and Tsoumakas, Grigorios},
  journal={Complex \& Intelligent Systems},
  volume={8},
  number={6},
  pages={4663--4678},
  year={2022},
  publisher={Springer}
}

@inproceedings{vidgen2021learning,
  title={Learning from the worst: Dynamically generated datasets to improve online hate detection},
  author={Vidgen, Bertie and Thrush, Tristan and Talat, Zeerak and Kiela, Douwe},
  booktitle={Proceedings of the 59th annual meeting of the Association for Computational Linguistics and the 11th international joint conference on natural language processing (volume 1: long papers)},
  pages={1667--1682},
  year={2021}
}

@article{alharthi2023target,
  title={Target-oriented investigation of online abusive attacks: A dataset and analysis},
  author={Alharthi, Raneem and Alharthi, Rajwa and Shekhar, Ravi and Zubiaga, Arkaitz},
  journal={IEEE Access},
  volume={11},
  pages={64114--64127},
  year={2023},
  publisher={IEEE}
}

@inproceedings{golbeck2017,
  title={A large labeled corpus for online harassment research},
  author={Golbeck, Jennifer and others},
  booktitle={Proceedings of the 2017 ACM on web science conference},
  year={2017}
}

@inproceedings{anzovino2018automatic,
  title={Automatic identification and classification of misogynistic language on twitter},
  author={Anzovino, Maria and Fersini, Elisabetta and Rosso, Paolo},
  booktitle={International Conference on Applications of Natural Language to Information Systems},
  pages={57--64},
  year={2018},
  organization={Springer}
}

@inproceedings{hammer2019,
  title={Threat: A large annotated corpus for detection of violent threats},
  author={Hammer, Hugo L and others},
  booktitle={2019 International Conference on Content-Based Multimedia Indexing (CBMI)},
  year={2019},
  organization={IEEE}
}

@inproceedings{hammer2014detecting,
  author    = {Hammer, Hugo Lewi},
  title     = {Detecting threats of violence in online discussions using bigrams of important words},
  booktitle = {2014 {IEEE} Joint Intelligence and Security Informatics Conference},
  year      = {2014},
  publisher = {{IEEE}}
}

@inproceedings{ravi2023exploring,
  author    = {Ravi, Kamalakkannan and others},
  title     = {Exploring multi-level threats in {Telegram} data with {AI}-human annotation: a preliminary study},
  booktitle = {2023 International Conference on Machine Learning and Applications (ICMLA)},
  year      = {2023},
  publisher = {IEEE},
  organization = {IEEE}
}

@article{bruni2025amaqa,
  title={Amaqa: A metadata-based qa dataset for rag systems},
  author={Bruni, Davide and Avvenuti, Marco and Tonellotto, Nicola and Tesconi, Maurizio},
  journal={arXiv preprint arXiv:2505.13557},
  year={2025}
}

@article{alvisi2025mapping,
  title={Mapping the italian telegram ecosystem: Communities, toxicity, and hate speech},
  author={Alvisi, Lorenzo and Tardelli, Serena and Tesconi, Maurizio},
  journal={arXiv preprint arXiv:2504.19594},
  year={2025}
}

@misc{laurer_modernbert_zeroshot_2024,
  author       = {Laurer, Moritz},
  title        = {ModernBERT-large-zeroshot-v2.0},
  year         = {2024},
  publisher    = {Hugging Face},
  howpublished = {\url{https://huggingface.co/MoritzLaurer/ModernBERT-large-zeroshot-v2.0}}
}

@inproceedings{wester2016threat,
  title={Threat detection in online discussions},
  author={Wester, Aksel and {\O}vrelid, Lilja and Velldal, Erik and Hammer, Hugo Lewi},
  booktitle={Proceedings of the 7th Workshop on Computational Approaches to Subjectivity, Sentiment and Social Media Analysis},
  pages={66--71},
  year={2016}
}

@article{roth2015context,
  title={Context-aware frame-semantic role labeling},
  author={Roth, Michael and Lapata, Mirella},
  journal={Transactions of the Association for Computational Linguistics},
  volume={3},
  pages={449--460},
  year={2015},
  publisher={MIT Press One Rogers Street, Cambridge, MA 02142-1209, USA journals-info~…}
}

\end{document}